\documentclass{article}

\usepackage[preprint,nonatbib]{neurips_2024}

\usepackage[utf8]{inputenc} 
\usepackage[T1]{fontenc}    
\usepackage[colorlinks=true,citecolor=brown,urlcolor=gray]{hyperref}

\usepackage{url}            
\usepackage{wrapfig,lipsum,booktabs}       
\usepackage{amsfonts}       
\usepackage{nicefrac}       
\usepackage{microtype}      
\usepackage[dvipsnames]{xcolor}  

\usepackage{colortbl}
\usepackage{amssymb}
\usepackage{pifont}
\usepackage{rotating}
\usepackage{makecell}
\usepackage{enumitem}

\usepackage{bbm}

\usepackage{caption}
\usepackage{subcaption}
\usepackage{amsmath}
\usepackage{xspace}
\usepackage{multirow}

\usepackage{algorithm}
\usepackage{algpseudocode}

\usepackage[titletoc]{appendix}

\usepackage{pdflscape}

\newcommand{\StartMenu}{\raisebox{-0.18cm}{\includegraphics[scale=0.022]{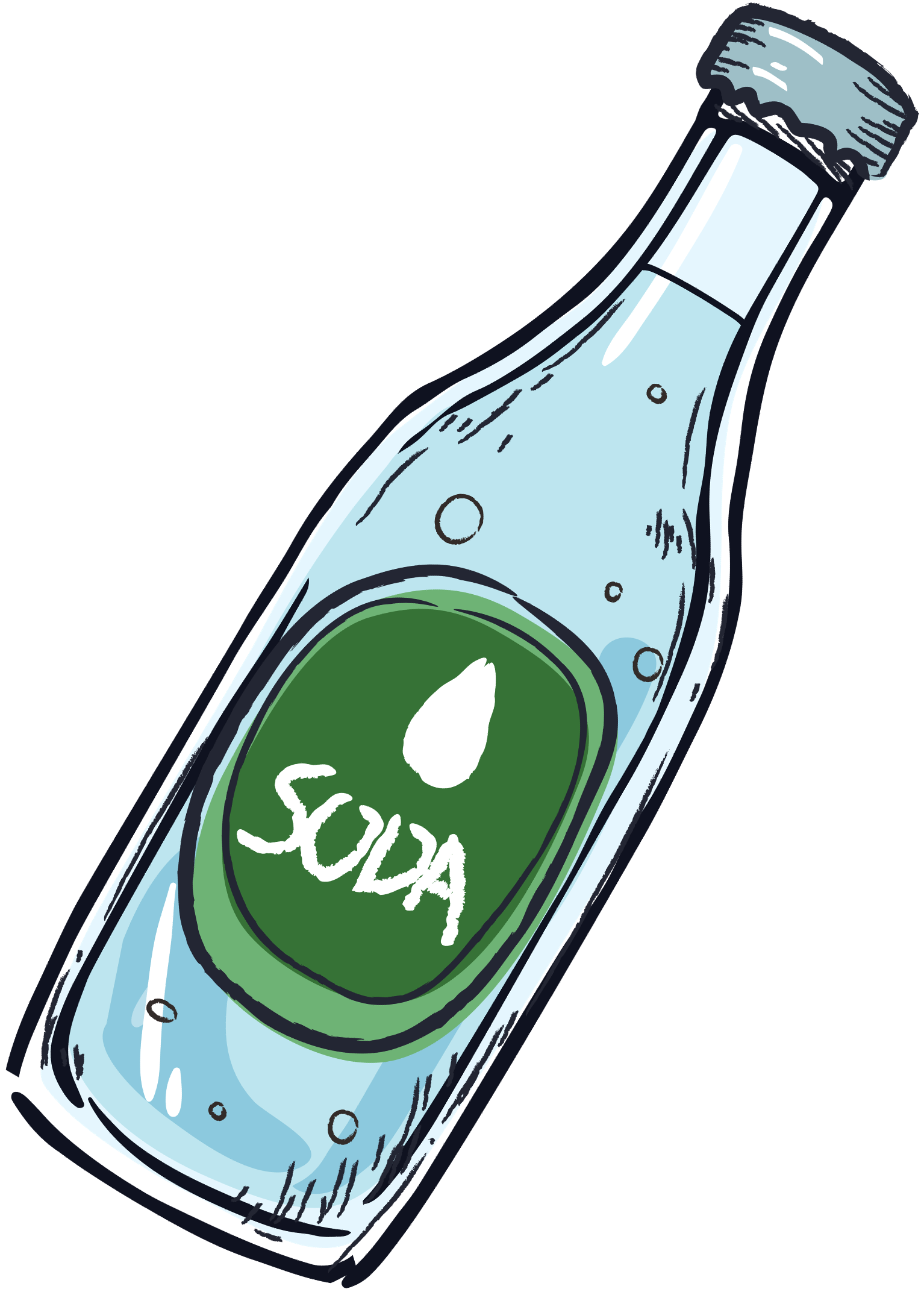}}}%
\newcommand{\SODA}[1]{%
   \StartMenu
   \foreach \x in {#1} {%
   \texttt{\textbf{\x}}%
   }%
}%

\definecolor{mygreen}{RGB}{0, 128, 0}
\definecolor{myred}{RGB}{255, 0, 0}
\definecolor{myblue}{RGB}{0, 0, 255}
\definecolor{myorange}{RGB}{255, 165, 0}
\definecolor{myblue2}{RGB}{224, 235, 246}
\definecolor{myred2}{RGB}{247, 230, 216}
\definecolor{myblack2}{RGB}{208, 206, 206}

\newlength\savewidth

\newcolumntype{C}{>{\centering\let\newline\\\arraybackslash\hspace{0pt}}m{2cm}}

\usepackage[capitalize]{cleveref}
\crefname{section}{Sec.}{Secs.}
\Crefname{section}{Section}{Sections}
\Crefname{table}{Table}{Tables}
\crefname{table}{Tab.}{Tabs.}

\usepackage{amsmath,amsfonts,bm,eqnarray}
\usepackage{cancel}
\usepackage{amsthm}
\usepackage{tikz}
\usetikzlibrary{tikzmark}

\newtheorem*{remark}{Remark}

\newtheorem{innercustomgeneric}{\customgenericname}
\providecommand{\customgenericname}{}
\newcommand{\newcustomtheorem}[2]{%
  \newenvironment{#1}[1]
  {%
   \renewcommand\customgenericname{#2}%
   \renewcommand\theinnercustomgeneric{##1}%
   \innercustomgeneric
  }
  {\endinnercustomgeneric}
}

\newcustomtheorem{customThm}{Theorem}
\newcustomtheorem{customLemma}{Lemma}
\newcustomtheorem{customCor}{Corollary}
\newcustomtheorem{customProposition}{Proposition}

\def\eqref#1{equation~\ref{#1}}

\def\1{\bm{1}}

\def\vh{{\bm{h}}}

\def\vu{{\bm{u}}}
\def\vv{{\bm{v}}}

\def\vx{{\bm{x}}}

\def\mA{{\bm{A}}}
\def\mB{{\bm{B}}}

\def\mI{{\bm{I}}}

\def\mL{{\bm{L}}}

\def\mQ{{\bm{Q}}}
\def\mR{{\bm{R}}}
\def\mS{{\bm{S}}}

\def\mU{{\bm{U}}}
\def\mV{{\bm{V}}}
\def\mW{{\bm{W}}}

\DeclareMathAlphabet{\mathsfit}{\encodingdefault}{\sfdefault}{m}{sl}
\SetMathAlphabet{\mathsfit}{bold}{\encodingdefault}{\sfdefault}{bx}{n}

\renewcommand{\frac}{\tfrac}

\newcommand{\bSigma}{{\boldsymbol{\Sigma}}}
\newcommand{\bsigma}{{\boldsymbol{\sigma}}}
\newcommand{\bdelta}{{\boldsymbol{\delta}}}

\title{\SODA{SODA}: Spectrum-Aware Parameter-Efficient Fine-Tuning for Diffusion Models}

\author{%
  \vspace{-5mm}\\
  \textbf{\small Xinxi Zhang\textsuperscript{1,*}~~~~Song Wen\textsuperscript{1,*}~~~~Ligong Han\textsuperscript{1,*,\textdagger}~~~~Felix Juefei-Xu\textsuperscript{2}~~~~Akash Srivastava\textsuperscript{3}}\\
  \textbf{Junzhou Huang\textsuperscript{4}~~~~
  \small Hao Wang\textsuperscript{1}~~~~Molei Tao\textsuperscript{5}~~~~Dimitris Metaxas\textsuperscript{1}}\\[0.5mm]
  \small \textsuperscript{1}Rutgers University~~~~\textsuperscript{2}New York University~~~~\textsuperscript{3}MIT-IBM Watson AI Lab~~~~\small \textsuperscript{4}UT Arlington~~~~\textsuperscript{5}Georgia Tech\\
  \small \textsuperscript{*}Equal contribution~~~~~\textsuperscript{\textdagger}Project lead, Corresponding author
}

\begin{document}

\maketitle
\begin{figure}[H]
\centering
\includegraphics[width=1\textwidth]{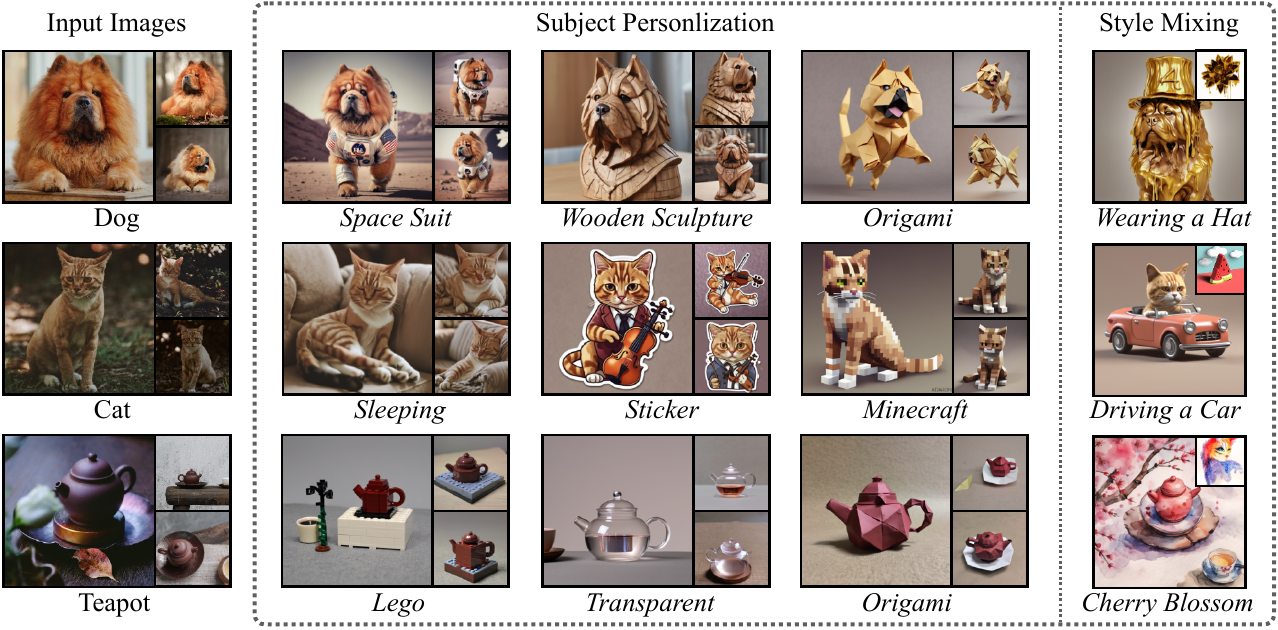} 
\caption{\textbf{SODA} achieves superior image quality and text alignment across diverse input images and prompts, such as changing the background, altering the texture, and synthesizing new poses. Additionally, SODA can generate prompt-aligned images in a given style specified by an input style image.} 
\label{fig.main} 
\end{figure}

\begin{abstract}
  Adapting large-scale pre-trained generative models in a parameter-efficient manner is gaining traction. Traditional methods like low rank adaptation achieve parameter efficiency by imposing constraints but may not be optimal for tasks requiring high representation capacity. We propose a novel spectrum-aware adaptation framework for generative models. Our method adjusts both singular values and their basis vectors of pretrained weights. Using the Kronecker product and efficient Stiefel optimizers, we achieve parameter-efficient adaptation of orthogonal matrices. Specifically, we introduce {\bf S}pectral {\bf O}rthogonal {\bf D}ecomposition {\bf A}daptation (SODA), which balances computational efficiency and representation capacity. Extensive evaluations on text-to-image diffusion models demonstrate SODA's effectiveness, offering a spectrum-aware alternative to existing fine-tuning methods.
\end{abstract}

\section{Introduction}\label{sec:intro}
Adapting large-scale pre-trained vision generative foundation models, such as Stable Diffusion~\cite{rombach2022high,esser2024scaling,blattmann2023stable}, in a parameter-efficient manner, is increasingly gaining traction within the research community. These generative models, which have demonstrated remarkable capabilities in generating high-quality images, can be computationally intensive and require substantial memory resources. To make these models more accessible and adaptable to various applications~\cite{kawar2023imagic,Zhang_2023_ICCV,black2023training,xue2024raphael}, researchers are focusing on methods that fine-tune these models efficiently without necessitating retraining the entire network. Parameter-efficient adaptation~\cite{mou2024t2i,ding2023parameter,sun2023simple} not only reduces computational overhead but also enables quicker and more flexible model deployment across different tasks and datasets.

The potential for parameter-efficient fine-tuning has been highlighted through extensive validations, demonstrating the ability to adapt base models to various data, enabling enhancements and customizations tailored to specific tasks and user characteristics. These methods allow the underlying model architecture to remain largely unchanged while inserting or adjusting a small subset of parameters. This approach is advantageous because it preserves the pre-trained knowledge while introducing task-specific adjustments. The lightweight nature of the optimized parameters also facilitates their seamless integration, making it possible to achieve high performance without the computational costs associated with full model retraining.

The efficiency of these methods is generally achieved by introducing structures or constraints into the parameter space. For instance, Low Rank Adaptation (LoRA)~\cite{hu2021lora} constrains parameter changes to be low-rank, while Kronecker Adapter (KronA)~\cite{edalati2022krona} constrains weight matrix changes to be a Kronecker product. By imposing these constraints, the adaptation process becomes more manageable and computationally efficient. LoRA, for example, operates under the assumption that the necessary adjustments to the model’s parameters are intrinsically low-dimensional, which simplifies the optimization process. However, this low-rank constraint might not always be optimal, especially for tasks requiring higher representation capacity, as it might limit the model’s ability to capture complex patterns in the data.

Despite the simplicity and effectiveness of LoRA, its low-rank constraint may not be optimal for tasks requiring high representation capacity. Specifically, for a rank $r$ approximation of a matrix $\mW$, the optimal solution corresponds to the largest $r$ singular values and their associated singular vectors, which LoRA does not explicitly utilize. This limitation suggests that there are potentially valuable directions in the parameter space, represented by the singular vectors, that are not being exploited. Recognizing this gap, we propose to leverage the full spectral information of the pretrained weight matrices during the fine-tuning process, thereby enhancing the model’s adaptability and performance.

In this paper, we propose a novel approach for spectrum-aware adaptation of generative models. Our method leverages the spectral space of pretrained weights, adjusting both singular values and singular vectors during fine-tuning. By focusing on both the magnitude and direction of these spectral components, we can achieve a more nuanced and effective adaptation. To ensure parameter efficiency, we employ a Kronecker product to rotate the singular vectors, thus modifying both their magnitude and direction. This approach allows us to maintain a balance between computational efficiency and the ability to capture complex data representations, making our method particularly suitable for high-dimensional tasks.
Our contributions are as follows:
\begin{itemize}[nosep]
    \item We propose a fine-tuning framework for personalization of text-to-image diffusion models by utilizing the spectrum of the pretrained parameters.
    \item We introduce SODA, {\bf S}pectral {\bf O}rthogonal {\bf D}ecomposition {\bf A}daptation, a parameter-efficient formulation of spectrum aware fine-tuning framework for generative models that leverages Kronecker product and jointly adjusts the magnitude and orientation of the parameter's singular vectors during fine-tuning.
    \item We conduct extensive evaluations for our method on customization of text-to-image diffusion models, and demonstrate it serves as an attractive alternative to traditional parameter-efficient fine-tuning methods that are not spectrum-aware.
\end{itemize}

\section{Related Work}
\label{sec:related}
\noindent\textbf{Diffusion personalization.} Diffusion~\cite{sohl2015deep,ho2020denoising,song2020score,saharia2022photorealistic,ramesh2022hierarchical,rombach2022high} personalization~\cite{gal2022image,ruiz2023dreambooth} aims to learn or reproduce concepts or subjects using pre-trained diffusion models, given one or a few images. Some works~\cite{ruiz2023dreambooth,kumari2023multi,lee2024direct} fine-tune the pre-trained diffusion models on images containing the desired concepts or subjects. DreamBooth~\cite{ruiz2023dreambooth} proposes fine-tuning the entire set of weights to represent the subjects or concepts as unique identifiers, which can be used for synthesizing images of different scenarios or styles. CustomDiffusion~\cite{kumari2023multi} suggests that fine-tuning only the cross-attention layers is sufficient to learn new concepts, leading to better performance on multiple-concept compositional generation. Lee et al. propose DCO~\cite{lee2024direct}, which fine-tunes the diffusion models without losing the composition ability of the pre-trained models by using implicit reward models. Another line of work~\cite{gal2022image,fei2023gradient} focuses on optimizing the word embeddings. Specifically, Textual Inversion~\cite{gal2022image} proposes optimizing word embeddings to capture unique concepts while freezing the pre-trained diffusion model weights. Additionally, some works~\cite{gal2023encoder,wei2023elite,avrahami2023break} combine the optimization of word embeddings and diffusion model weights. However, model fine-tuning methods typically involve a large number of parameters, which can be inefficient and prone to overfitting.


\noindent\textbf{Parameter efficient fine-tuning.} The rapid development of foundation models, which contain a large number of parameters, has made fine-tuning these models on small datasets challenging due to the numerous parameters involved. To address the efficiency and overfitting issues in model fine-tuning, Parameter Efficient Fine-Tuning (PEFT) techniques have been proposed. One line of PEFT research focuses on Adapter tuning~\cite{rebuffi2017learning,hu2023llm,chen2022adaptformer}, which involves inserting trainable layers within the layers of pre-trained models. Another line of research explores Low-Rank Adaptation (LoRA)~\cite{hu2021lora,zhang2023adaptive,dettmers2024qlora,kopiczko2023vera,zhu2024asymmetry,hayou2024lora+}. LoRA~\cite{hu2021lora,lora_stable} proposes learning residual weights by constructing them through the multiplication of two low-rank matrices, thereby significantly reducing the number of learned parameters. Other methods have also been developed, such as SVDiff~\cite{han2023svdiff}, which performs singular value decomposition on the pre-trained weight matrices and only fine-tunes the singular values, and OFT~\cite{qiu2023controlling,liu2023parameter}, which maintains the hyperspherical energy of the pre-trained model by multiplying a trainable orthogonal matrix. Additionally, KronA~\cite{edalati2022krona,marjit2024diffusekrona} constructs the residual weight matrices using the Kronecker product of two small-size matrices. However, the aforementioned methods do not fully leverage the prior knowledge embedded in the pre-trained weights. In this work, we enhance existing PEFT methods by introducing Spectral Orthogonal Decomposition Adaptation. While a concurrent study~\cite{zhang2024spectral} also employs a spectrum-aware approach, it differs from ours as it solely focuses on fine-tuning the top spectral space.

\section{Methodology}
\label{sec:method}
\begin{figure}[t]
\centering
\includegraphics[width=1\textwidth]{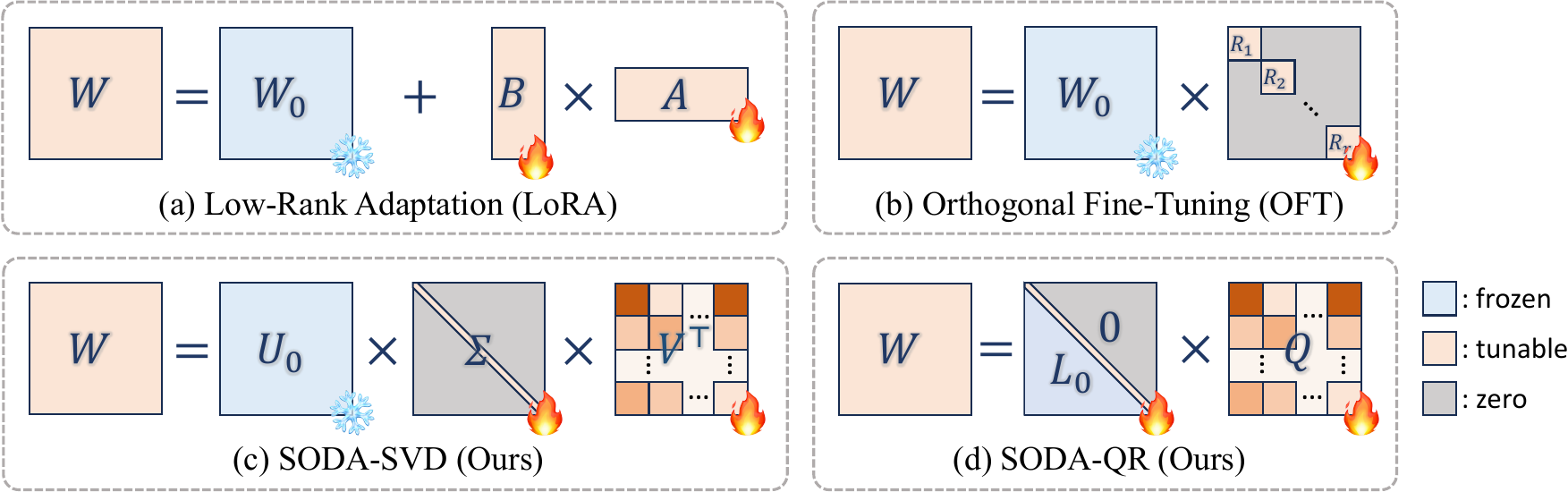} 
\caption{Comparison of difference PEFT approaches. (\textcolor{myblue2}{\ding{110}}: frozen parameters; \textcolor{myred2}{\ding{110}}: tunable parameters; \textcolor{myblack2}{\ding{110}}: zeros.)}
\label{fig.compare} 
\end{figure}

\subsection{Preliminary}
\noindent \textbf{Low-Rank Adaptation (LoRA).}
Text-to-image diffusion models consist of numerous large pre-trained weights. We follow LoRA's Stable Diffusion implementation~\cite{lora_stable} and only fine-tune the linear projection matrices in cross-attention layers, one of which is denoted as $\mW_0 \in \mathbb{R}^{m\times n}$. The weight change during the fine-tuning process is denoted as $\Delta \mW$. Low-Rank Adaptation~(LoRA) assumes the low rank of the network's weight increments and decomposes each increment matrix into the product of two low-rank matrices $\Delta \mW = \mB \mA$, where $ \mA \in \mathbb{R}^{r \times n} $ and $\mB \in \mathbb{R}^{m \times r} $. Therefore, we can derive the following forward propagation: 
\begin{equation}
    \vh = \mW_0 \vx + \Delta \mW \vx = \mW_0 \vx + \mB \mA \vx, 
\end{equation}
where $\vh$ and $\vx$ respectively represent the output and input of $\mW$. 

\noindent \textbf{Orthogonal Fine-Tuning (OFT).} 
Finetuning diffusion models usually requires efficiency and prior knowledge preservation. To enhance prior knowledge, OFT~\cite{qiu2023controlling} proposes to retain hyperspherical energy in pairwise relational structure among neurons. In detail, it learns an orthogonal matrix to conduct the same transformation for all neurons in each layer, which keeps the angle among all neurons in each layer unchanged. The weight update is represented by
\begin{equation}
\mW = \mW_0 \mR,
\end{equation}
where $\mR$ is an orthogonal matrix and $\mW_0$ is the pre-trained weight. To decrease the number of trainable parameters, the original OFT uses a block-diagonal structure to make it parameter efficient, where $\mR := \text{diag}(\mR_1,\mR_2,\ldots,\mR_r)$ and each $\mR_i$ is a small-size orthogonal matrix.

\subsection{Optimization on Stiefel Manifold}
A Stiefel manifold $St(n,m):= \{\mV \in \mathbb{R}^{n \times m}: \mV^{\top}\mV=\mI_{m\times m}\}$ is the set of $n \times m$ matrices ($n \le m$) with each column orthogonal to all other columns. Consider the optimization problem $min_{\mV \in St(n,m)}f(\mV)$, which is to find a matrix $\mV$ that minimizes a given objective function $f(\mV)$ subject to the constraint that $\mV$ lies on a Stiefel manifold. This optimization problem has plenty of applications, such as the OFT above.
To make sure that the parameter being optimized stays on the Stiefel manifold, the original OFT use Cayley parameterization, $\mR=(\mI+\mS)(\mI-\mS)^{-1}$ where $\mS$ is a skew-symmetric matrix. Then $\mR$ is an orthogonal matrix and OFT only needs to optimize $\mS$. Here we utilize the Stiefel optimizer introduced in \cite{kong2022momentum}, which preserves the manifold structure and keeps momentum in the cotangent space.
Given an orthogonal matrix $\mV$, we can use a Stiefel optimizer to keep it in the Stiefel manifold. However, directly optimizing $\mV$ is not parameter efficient. Here we utilize Kronecker product, we leverage the following remark:
\begin{remark}
    If $\mV_1, \mV_2, \ldots, \mV_r$ are orthogonal matrices, then their Kronecker product $\bigotimes_{i=1}^r{\mV_i}=\mV_1 \otimes \mV_2 \otimes \cdots \otimes \mV_r$ is also orthogonal.
\end{remark}
Then, the Kronecker product of several small-size orthogonal matrices can generate a relatively large-size orthogonal matrix, so the number of parameters is reduced. Such formulation is more efficient than OFT, we call it {\bf K}ronecker {\bf O}rthogonal {\bf F}ine-{\bf T}uning (KOFT).
However, it is not always possible to find a Kronecker decomposition for any given orthogonal matrix. Thus, inspired by OFT, we learn a rotation matrix $\mR$ to adjust $\mV_0$, $\mV_\mR := \mV_0 \mR$. Now, since $\mR$ is initialized as identity $\mI$, we can always parameterize it using Kronecker product, $\mV_\mR=\mV_0 (\mR_1 \otimes \mR_2)$. The shared block diagonal structure as adopted by the original OFT corresponds to $\mI_{r \times r}\otimes\mR$. Comparing with ours, this is more sparse.
So why do we want to consider optimizing in Stiefel manifold? On the one hand, orthogonal matrices naturally arise in numerical decomposition. On the other hand, previous works~\cite{achour2022existence,bansal2018can} have shown that imposing orthogonality on model parameters facilitates the learning by limiting the exploding/vanishing gradients and improves the robustness.

\subsection{Spectrum Aware Fine-Tuning}

LoRA and OFT have shown promising performance in adapting pre-trained models to downstream tasks. However, LoRA neglects the prior knowledge in the pre-trained weights, while OFT only utilizes the angle information among neurons, failing to explore the knowledge in the pre-trained weights fully.
To better utilize the spectrum of pre-trained weights, we propose {\bf S}pectral {\bf O}rthogonal {\bf D}ecomposition {\bf A}daptation (SODA). We first decompose the pre-trained weight matrix $\mW_0$ in each layer into a spectral component $\mW_0^{spec}$ and a basis matrix $\mW_0^{basis}$, formulated as $\mW_0 = \mW_0^{spec}\mW_0^{basis}$. We then update the spectrum in the spectral component $\mW_0^{spec}$ and the basis matrix $\mW_0^{basis}$ separately. The spectrum in the spectral component $\mW_0^{spec}$ is optimized using a gradient descent optimizer. As the basis matrix is always orthogonal, we could use a Stiefel optimizer to optimize it on the Stiefel manifold. However, directly optimizing $\mW_0^{basis}$ would result in a number of trainable parameters as large as full-weight tuning. Considering that both the matrix product and Kronecker product of orthogonal matrices are also orthogonal, we construct an orthogonal matrix $\mR = \bigotimes_{i=1}^r{\mR_i}$, where $\mR_i$ is a small-size orthogonal matrix.
Then our updated weight can be formulated as:
\begin{align}
\mW = (\mW_0^{spec} \oplus \underline{\Delta{\mS}}) \mW_0^{basis} \underline{{\mR}},
\end{align}
where the parameters with underlines are trainable. $\Delta\mS$ denotes the incremental spectrum, where the operator $\oplus$ represents the addition of the incremental spectrum to the spectrum in the spectral component matrix $\mW_0^{spec}$. The incremental spectrum $\Delta\mS$ is updated using a gradient descent optimizer, while each orthogonal matrix $\mR_i$ is updated using a Stiefel optimizer, which ensures that the orthogonality constraint is maintained during the optimization process. 
Here we consider two decomposition methods, SVD and LQ/QR decomposition:

\noindent \textbf{Singular Value Decomposition (SVD).}
If we decompose $\mW_0=\mU_0\mathbf{\Sigma_0}\mV_0^\top$ where $\bSigma_0=\text{diag}(\bsigma)$ is singular values and $\mV_0^\top$ is an orthogonal matrix, SVDiff~\cite{han2023svdiff} proposes to fine-tune the singular values, or tuning the spectral shifts $\bdelta$,
\begin{equation}
\mW = \mU_0 \bSigma_\bdelta \mV_0^\top ~~ \text{with} ~~ \bSigma_\bdelta := \text{diag}(\text{ReLU}(\bsigma+\underline{\bdelta})).
\end{equation}
We propose to fine-tune the singular vectors $\mV$ as well. However, directly tuning $\mV$ would not be parameter-efficient. We thus leverage Kronecker product $\mV_{\mR} := \mV_0 \mR = \mV_0 (\bigotimes_{i=1}^r{\mR_i})$,
\begin{equation}
\mW_\text{SODA-SVD} = \mU_0 \bSigma_\bdelta \mV_\mR^\top ~~ \text{with} ~~ \bSigma_{\bdelta}=\text{diag}(\text{ReLU}(\bsigma+\underline{\bdelta})),\mV_{\mR} :=  \mV_0 (\bigotimes_{i=1}^r{\underline{\mR_i}}).
\end{equation}


\noindent \textbf{LQ/QR Decomposition (QR).}
We can alternatively decompose $\mW_0=\mL_0\mQ_0$, where $\mL_0$ is a lower triangle matrix and $\mQ_0$ is an orthonormal matrix. Similar to SVD, the diagonal of $\mL_0$ are eigenvalues of $\mL_0$ and we propose to fine-tune both $\mL_0$ and $\mQ_0$,
\begin{equation}
\mW_\text{SODA-QR} = \mL_\bdelta \mQ_\mR ~~ \text{with} ~~ \mL_\bdelta := \mL_0 + \text{diag}(\underline{\bdelta}), ~~\mQ_\mR := \mQ_0 (\bigotimes_{i=1}^r{\underline{\mR_i}}).
\end{equation}

\subsection{Analysis}
\noindent \textbf{Number of parameters.} A comparison of number of tunable parameters for different approaches is shown in Table~\ref{tab:numparam}. To simplify notation, here we assume $m=n$ and small rotation blocks are evenly divided ($n/r$ for OFT, and $n^{1/r}$ for SODA). For KOFT and SODA, we can observe that the parameter count decreases drastically as $r$ grows (we use $r=3$ in our experiments).
\begin{table}[b]
\caption{Comparison of Parameter Counts for Different Methods.}
\begin{center}
\setlength{\tabcolsep}{8pt}
\begin{tabular}{lcccc}
	\toprule
Method & LoRA & OFT & KOFT & SODA \\
\hline
\rule{0pt}{2.7ex}Number of parameters & $2 ~n \cdot r$ & $\frac{n^2}{r}$ \text{ or } $\frac{n^2}{r^2}$ \text{(shared)} & $r \cdot n^{2/r}$ & $n + r \cdot n^{2/r}$ \\
\bottomrule
\end{tabular}
\end{center}
\label{tab:numparam}
\end{table}

\noindent \textbf{Gradient of singular values.}
Given a matrix $\mW \in \mathbb{R}^{m \times n}$ and its singular value decomposition $\mW=\mU\bSigma\mV^\top$. If we denote the derivative of loss $l$ w.r.t. output $\vh$ as $\delta\vh=\frac{\partial l}{\partial\vh}$, then the gradient of $\mW$ equals to $\delta\mW=\delta\vh \vx^\top$, and
\begin{align}
    \delta\bSigma = \mU^\top(\delta\vh \vx^\top) \mV = (\mU^\top\delta\vh) (\mV^\top\vx)^\top, ~~\text{and}~~ \delta\bSigma_{ii} = \langle\vu_i, \delta\vh\rangle \langle\vv_i, \vx\rangle,
\end{align}
\noindent where $\langle\cdot,\cdot\rangle$ denotes inner product, and $\vu_i$ and $\vv_i$ are the $i$-th column of $\mU$ and $\mV$, respectively.
As can be seen, the gradient of singular values is composed of two parts: \textbf{a)} ``from left to right'', the gradient $\delta\vh$ is projected onto columns of $\mU$; \textbf{b)} ``from right to left'', the input $\vx$ is projected onto columns of $\mV$ and only the $i$-th component influences $\sigma_i$.

If we consider the change of weight matrix after a small step as $\Delta\mW$, then if we are only tuning singular values $\bsigma$, we can compute the effective change of the weight matrix $\Delta\mW'$,
\begin{align}
    \Delta\mW' &= \mU \Delta\bSigma \mV^\top\qquad\qquad\text{and}\\
    \Delta\bSigma &= (\mU^\top \Delta\mW \mV)\odot \mI_{m \times n}.
\end{align}
\noindent We can observe that
\begin{align}
    ||\Delta\mW'||_F^2 = ||\Delta\bSigma||_F^2 = ||(\mU^\top \Delta\mW \mV)\odot \mI_{m \times n}||_F^2 \leq ||\mU^\top \Delta\mW \mV||_F^2 = ||\Delta\mW||_F^2.
\end{align}
\noindent The $\leq$ sign comes from the fact that the Hadamard product $\odot$ masks out non-diagonal elements thus shrinks the Frobenius norm. In fact, since we will mask out most of the elements, $||\Delta\mW'||_F^2$ tends to be much smaller than $||\Delta\mW||_F^2$ and this is why we set a large learning rate to $\bsigma$.

\section{Experiments}
\label{sec:experiments}
We use Stable Diffusion XL (SDXL) \cite{podell2023sdxl} as the pre-trained T2I diffusion model. We conduct experiments on subject personalization (Sec. \ref{sec:subject_personalization}), style personalization (Sec.~\ref{sec:subject_personalization}), and ablation studies (Sec. \ref{ablation}). In all experiments, we train the text encoders and UNet of the SDXL model by replacing all linear modules in the attention and cross-attention layers with corresponding PEFT.

\subsection{Subject Personalization}
\label{sec:subject_personalization}

\begin{figure}[t]
\centering
\includegraphics[width=1\textwidth]{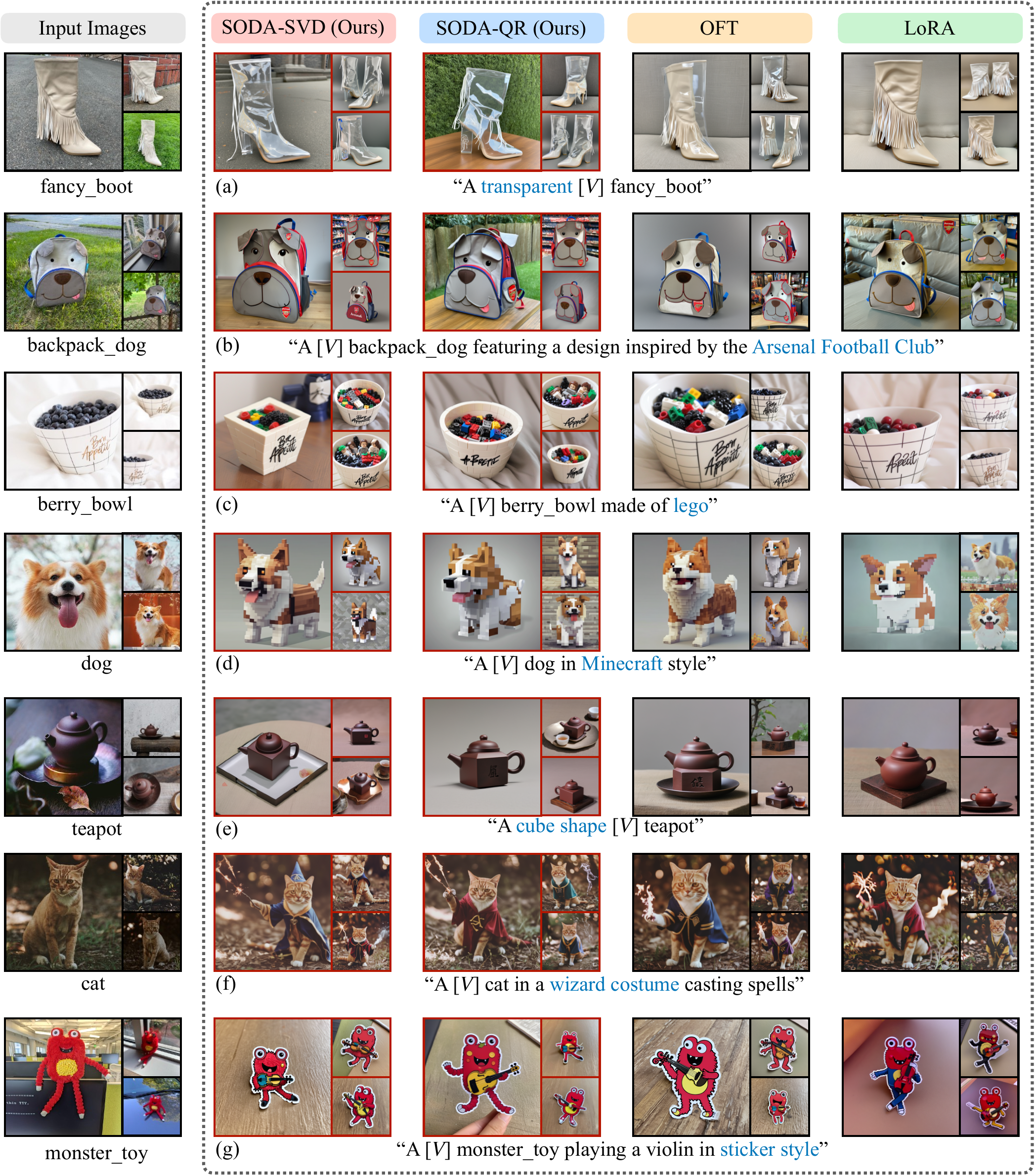} 
\caption{Results for \textbf{Subject Personalization}. Each subfigure consists of 3 samples: a large one on the left and two smaller ones on the right. The text under the input images indicates the class of the personalized subject, while the text prompt under the sample images is used for inference. Our observations indicate that SODA outperforms both LoRA and OFT in generating prompt-aligned images while preserving subject identities at a similar level.} 
\label{fig.subject} 

\end{figure}
\begin{figure}[h]
\centering
\includegraphics[width=1\textwidth]{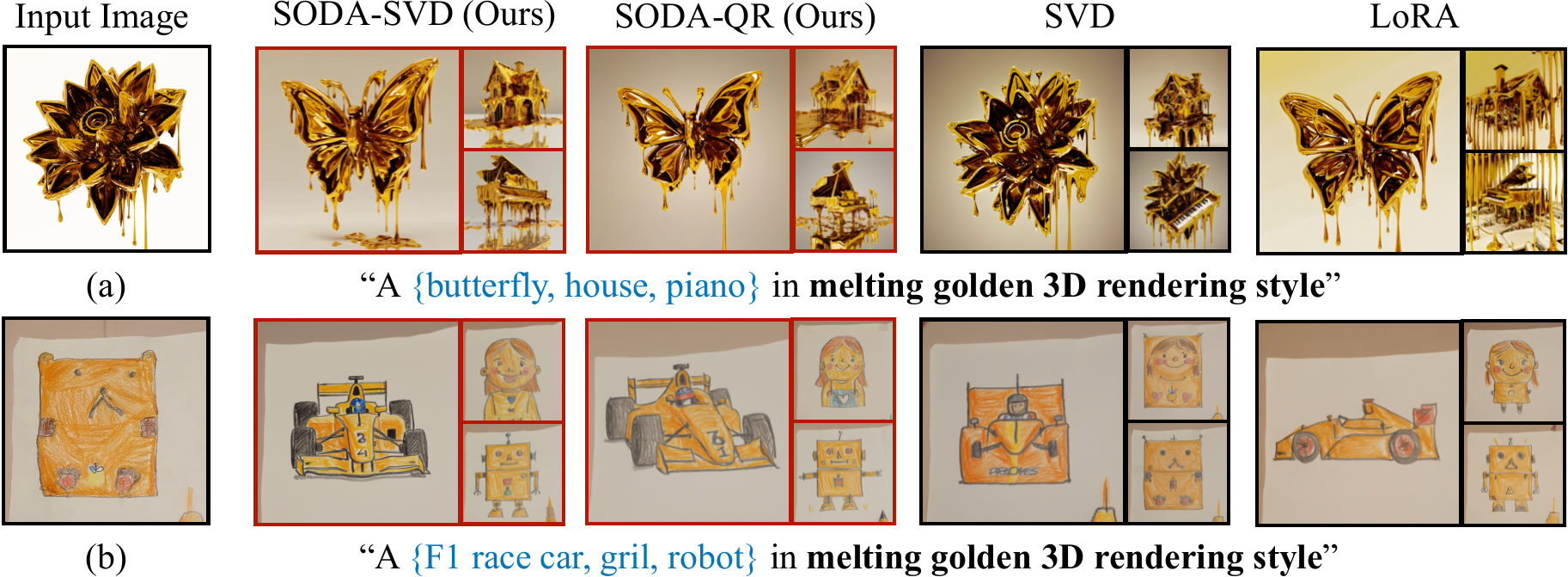} 
\caption{\textbf{Personalized style generation.} We show curated samples of ours (SODA-SVD, SODA-QR), SVDiff~\cite{han2023svdiff} (SVD), and LoRA~\cite{hu2021lora}. Independently trained subject and style weights are merged without joint training. SVDiff tends to overfit to the subject and fail to preserve the style well.} 
\label{fig.style} 
\end{figure}

\begin{figure}[ht]
\centering
\includegraphics[width=1\textwidth]{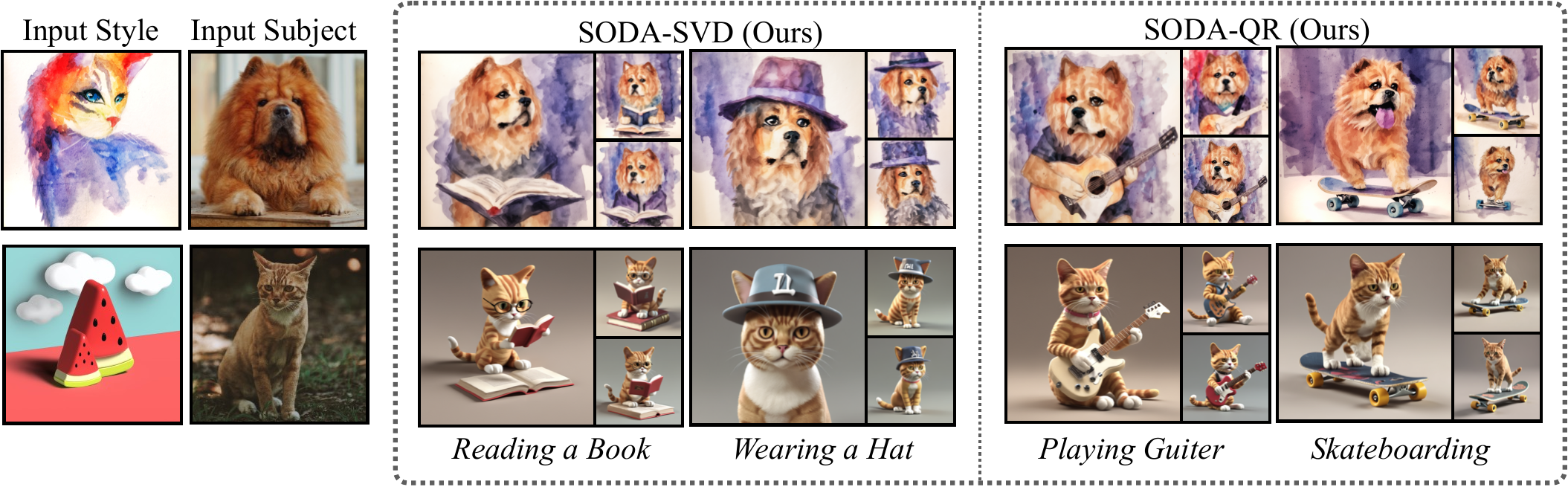} 
\caption{\textbf{Compositional generation of my subject in my style.} We show visual samples of {\em my subject in my style} with different actions or visual attributes specified by the text prompts. Independently trained subject and style weights are merged without joint training.} 
\label{fig.style_compositional} 
\end{figure}

\begin{wrapfigure}{r}{0.5\textwidth}
    \centering
    \vspace{-0em}
    \includegraphics[width=0.5\textwidth, trim=0 0 0 0, clip]{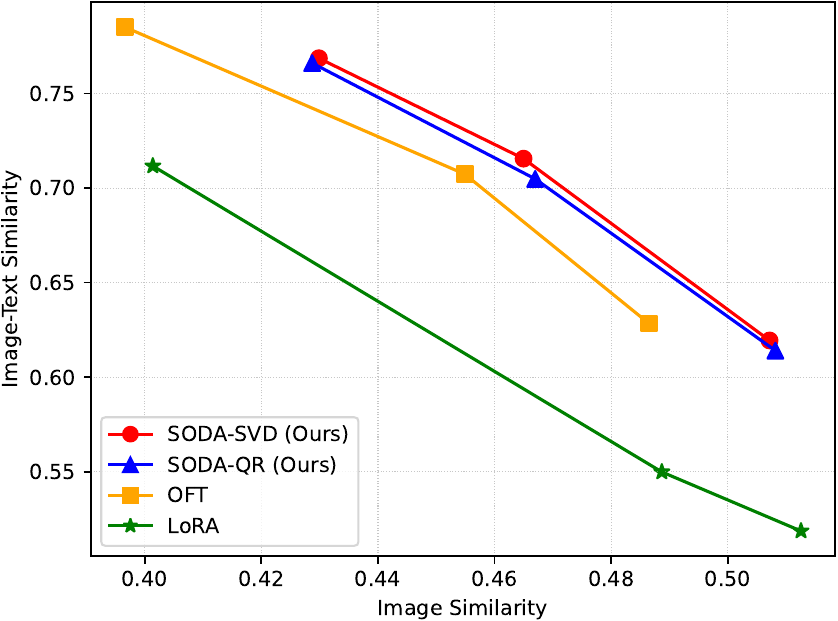}
    \caption{Pareto curve between subject fidelity (image similarity) and compositionality (image-text similarity) on subject personalization task of T2I diffusion models. Scores of each point on the curve are measured with {\em different learning rates}. Top-right corner is preferred.} 
    \label{fig.dino_clip}
    \vspace{-2.em}
\end{wrapfigure}

\textbf{Experimental setting.} 
For Subject Personalization, we fine-tuned the SDXL model on the DreamBooth dataset \cite{ruiz2023dreambooth} following the Direct Consistency Optimization (DCO) framework \cite{lee2024direct}. 
For a fair comparison, we tuned the best learning rate for each method and tested each method with three different learning rates. This allows us to evaluate their performance comprehensively.
Detailed experiment setting can be found in Appendix.

\textbf{Baselines.}
We compared our methods with strong baselines including LoRA~\cite{lora_stable} and OFT~\cite{qiu2023controlling}. For a fair comparison and to keep the number of parameters approximately the same, we set rank $r=1$ for LoRA and $r=3$ for KOFT and SODA, i.e., $\mR=\mR_1\otimes\mR_2\otimes\mR_3$. 

\textbf{Quantitative results.}
We report Image-Text Similarity ($\uparrow$, using SigLIP \cite{zhai2023sigmoid}) to measure the fidelity and Image Similarity ($\uparrow$, using DINOv2 \cite{oquab2023dinov2}) to measure the faithfulness or identity preservation. Detailed information on the evaluation prompts and metrics can be found in Appendix. We plot the Pareto curve consists of scores at {\em varying learning rates}. This curve illustrates the trade-off between the fidelity and faithfulness for the evaluated method. The upper right of the curve is ideal, indicating that the method can achieve prompt-aligned compositional generation while preserving the subject's identity.

Fig. \ref{fig.dino_clip} shows comparison of LoRA (\textcolor{mygreen}{\ding{72}}), our methods, SODA-SVD (\textcolor{myred}{\ding{108}}) and SODA-QR (\textcolor{myblue}{\ding{115}}). Interestingly, LoRA cannot push the frontier to the upper right, indicating that LoRA tends to overfit to the subject and struggles to generate prompt-aligned images while preserving the subject identity. This suggests that jointly adjusting the magnitude and orientation of the decomposed pretrained weight can better utilize model priors when adapting to new concepts without overfitting. Compared to OFT (\textcolor{myorange}{\ding{110}}), our methods SODA-SVD (\textcolor{myred}{\ding{108}}) and SODA-QR (\textcolor{myblue}{\ding{115}}) depict the upper-right frontier in both image-text similarity and image similarity, demonstrating their effectiveness. This suggests incorporating spectral tuning can further enhance performance. Interestingly, our methods, SODA-SVD (\textcolor{myred}{\ding{108}}) and SODA-QR (\textcolor{myblue}{\ding{115}}), overlap with each other, suggesting that tuning $\mathbf\Sigma$ has a very similar effect to tuning the diagonal of $\mathbf R$. 

\textbf{Qualitative results.}
In Fig. \ref{fig.subject}, we provide qualitative comparisons between our approaches (SODA-SVD and SODA-QR) and the baselines (OFT and LoRA). We observe that prompts involving background changes (f) or style changes (g) are handled well by all methods. These prompts are likely easier because they don't require a deep understanding of the subject; even overfitting can still produce prompt-aligned images. However, for prompts requiring changes in shape (e) or texture (a-d), LoRA struggles with compositional generation due to overfitting the object. Our methods and OFT perform significantly better than LoRA on prompts requiring texture or shape changes. For example, in (a), (c), and (e), our methods demonstrate superior compositional generation. This suggests the benefit of leveraging the spectrum of the pretrained weights, not just adjusting the basis orthogonally. These observations are also supported by the quantitative results discussed above. Interestingly, with objects like dogs (d), LoRA sometimes succeeds in generating good compositional images that changes the subject's texture. We hypothesize this is because pretrained models have a strong prior for such objects, making it easier for LoRA to find optimization points that represent the object well without overfitting.

\begin{wrapfigure}{l}{0.5\textwidth}
    \centering
    \vspace{-1.2em}
    \includegraphics[width=0.5\textwidth, trim=0 0 0 0, clip]{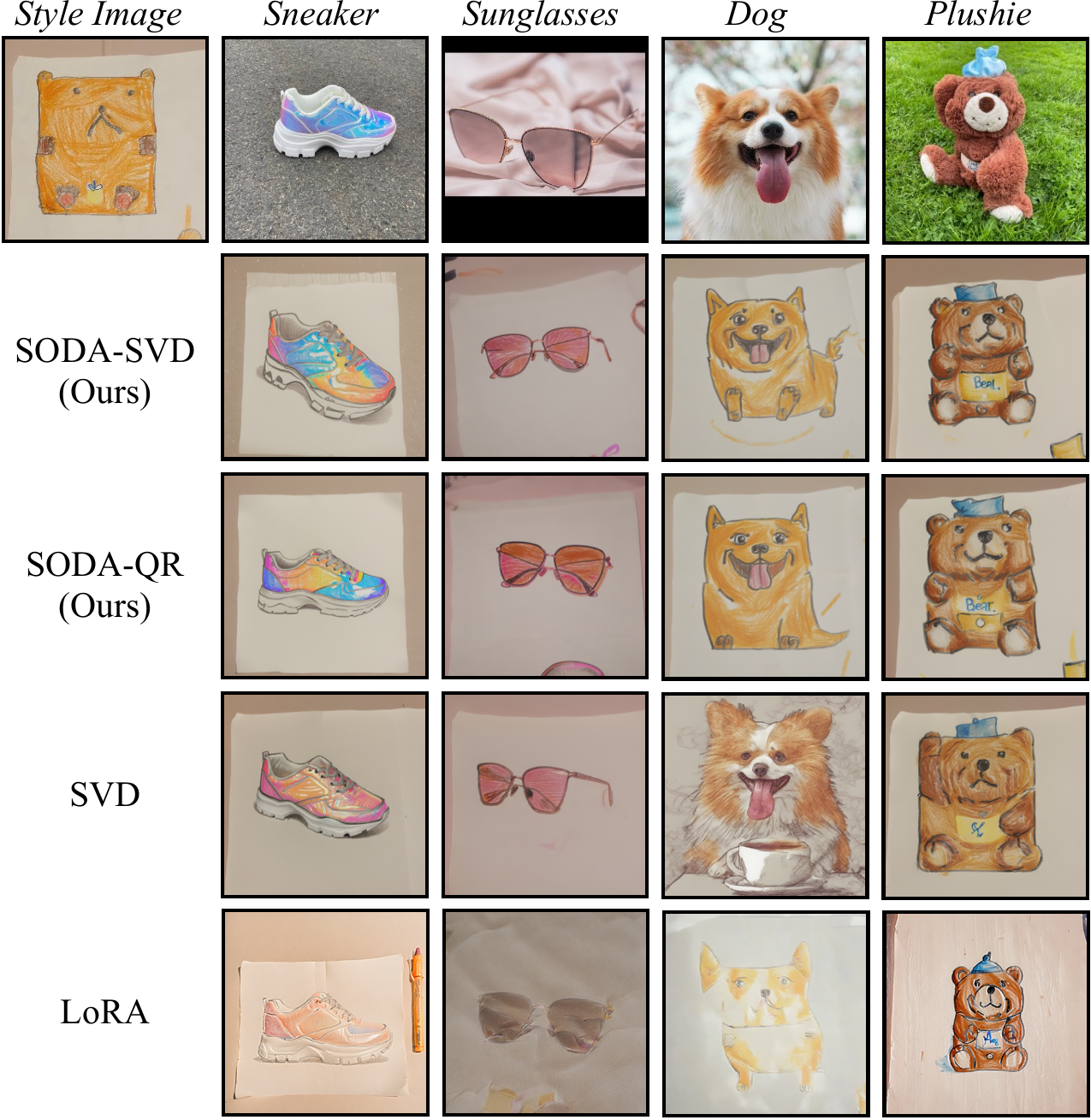}
    \caption{Comparison of my subject in my style.} 
    \label{fig.mixstyle}
    \vspace{-1em}
\end{wrapfigure}

\subsection{Style Personalization}
\label{sec:style_personalization}
\textbf{Experimental setting.} 
For style personalization we experiment on style images from StyleDrop dataset \cite{sohn2024styledrop}, we finetune all peft methods on 10 style reference images and generate comopositional images for the corresponded style. This result is displayed in Fig. \ref{fig.style}, And we follow \cite{sohn2024styledrop} to mix the personalized subject and style model to generate images of a personalized subject in a personalized style. we randomly picked 10 subjects from Dreambooth dataset \cite{ruiz2023dreambooth} and merge their residual fine-tuned peft weight $\Delta \mW_1$ with the residual fine-tuned style peft weight $\Delta \mW_2$. For merging, we use an arithmetic merge (Merge) \cite{sohn2024styledrop}, i.e., $\Delta \mW = \Delta \mW_1 + \Delta \mW_2$, result is displayed in Fig.~\ref{fig.mixstyle}. Other experiment settings follows Sec.~\ref{sec:subject_personalization}.

\textbf{Baselines.}
We compared our methods with LoRA with rank $r=1$, and SVDiff \cite{han2023svdiff}. For LoRA and SVDiff, we can directly get the merged residual weight by merging the residual weight of the subject personalization model and the style model. For our Methods, we calculate the residual weight by subtracting the fine-tuned SVD or QR weight by the pretrained weight and merge the residual weight. 

\textbf{Results.}
Fig. \ref{fig.style} shows results comparisons between LoRA, SVDiff and Ours (SODA-SVD and SODA-QR). Our methods can generate prompt-aligned images in the reference style, while SVDiff observed over-fit to the images, and LoRA generates artifacts.  Fig. \ref{fig.mixstyle} shows results of generated images by merging the subject model and the style model. Our models can generate style match images while preserving the identity of the personalized subject. while SVDiff tends to overfit to the subject and fail to preserve the style well, and lora also overfit to the subject and generate artifact. 
Additionally, we show novel compositional generation of the combined subject and style in Fig. \ref{fig.style_compositional}

\subsection{Ablation Study}
We also conduct ablation studies on spectral awareness and optimization on Stiefel Manifold to validate our design and choice.

\begin{figure}[ht]
    \centering
    \begin{subfigure}[b]{0.32\textwidth}
        \centering
        \includegraphics[width=\textwidth]{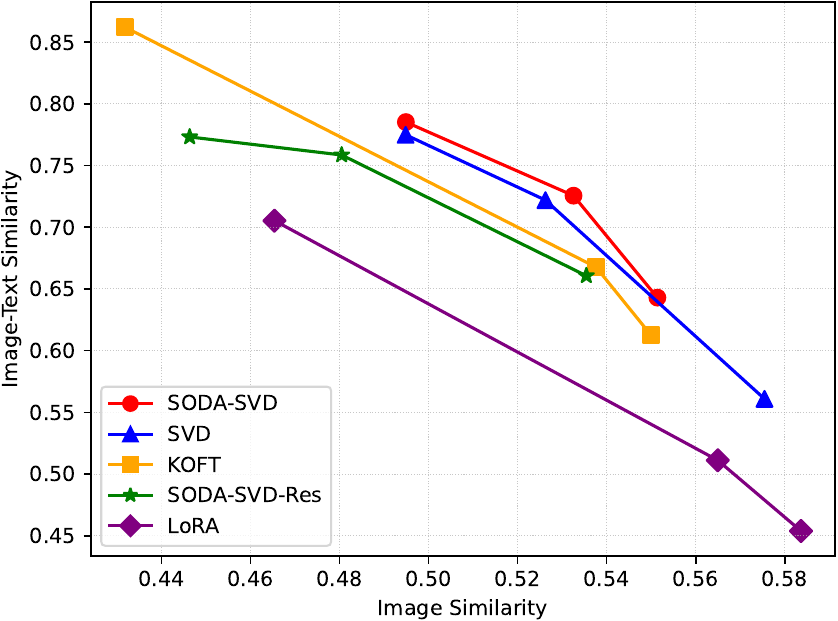}
        \caption{Effect of spectral and orthogonal tuning.}
        \label{fig:dino_clip1}
    \end{subfigure}
    \hfill
    \begin{subfigure}[b]{0.32\textwidth}
        \centering
        \includegraphics[width=\textwidth]{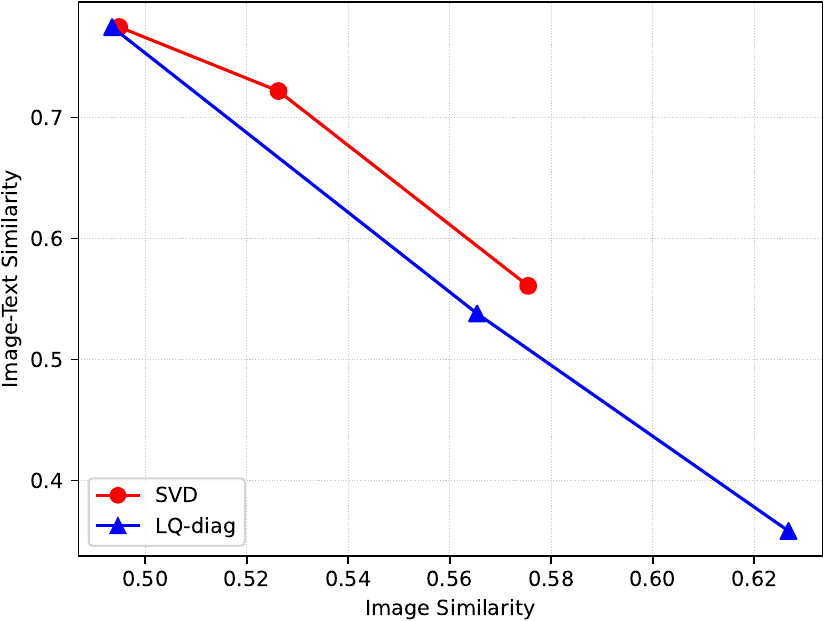}
        \caption{Choice of spectrum tuning strategies.}
        \label{fig:dino_clip2}
    \end{subfigure}
    \hfill
    \begin{subfigure}[b]{0.32\textwidth}
        \centering
        \includegraphics[width=\textwidth]{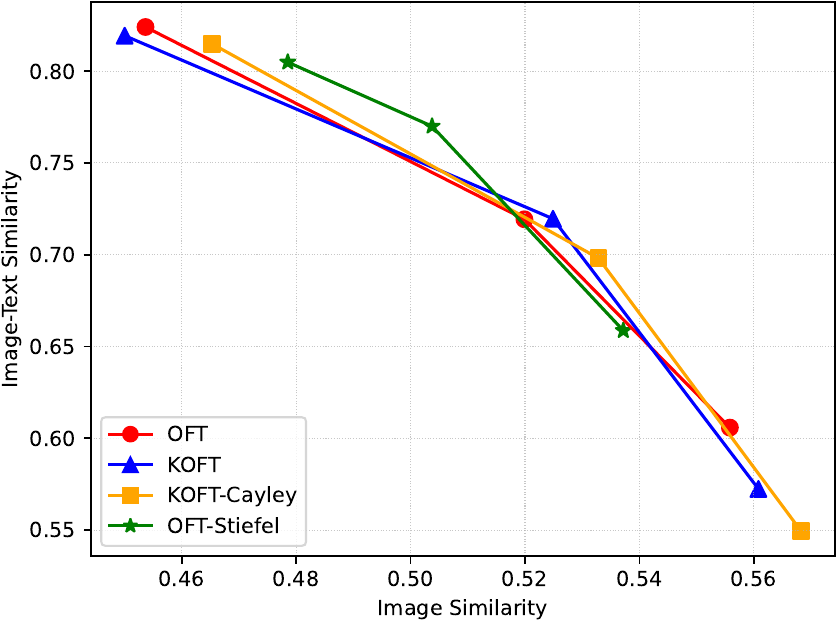}
        \caption{Choice of optimization on Siefel Manifold.}
        \label{fig:dino_clip3}
    \end{subfigure}
    \caption{Ablation studies on spectral awareness and optimization on Stiefel Manifold.}
    \label{fig:three_images}
\end{figure}
\label{ablation}
\begin{wrapfigure}{r}{0.5\textwidth}
    \centering
    \vspace{-1.2em}
    \includegraphics[width=0.5\textwidth]{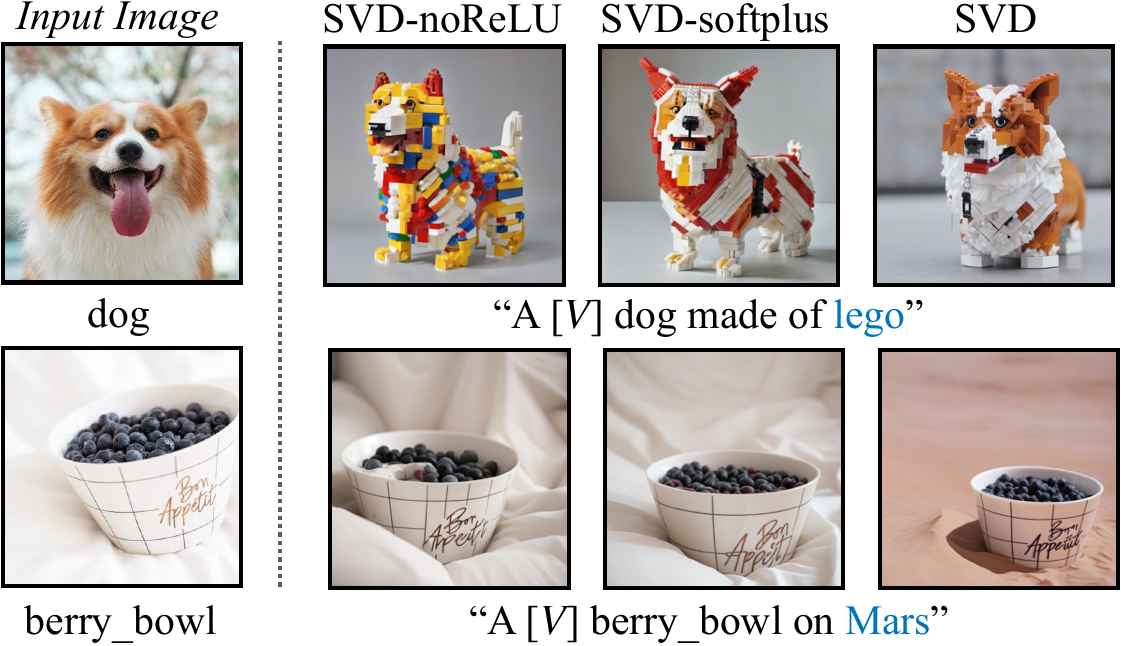}
    \caption{Results for spectral tuning on SVD.} 
    \label{fig.dino_clip4}
    \vspace{-1.em}
\end{wrapfigure}
\textbf{Spectral Awareness.}
\textbf{a)} We study the effect of spectral tuning and orthogonal tuning. We select 10 subjects from the DreamBooth dataset~\cite{ruiz2023dreambooth} and compare our method (SODA-SVD), which combines spectral tuning and orthogonal tuning, and only spectral tuning (SVD), only orthogonal tuning (Kronecker orthogonal Adapter), and a residual version of SODA-SVD. Fig.~\ref{fig:dino_clip1} shows orthogonal only and spectral only have similar performance to each other while ours performs better.
\textbf{b)} We also study the choice of spectral tuning method. In Fig.~\ref{fig:dino_clip2}, we compare the performance of spectral tuning using SVD and LQ/QR decomposition. The results demonstrate that SVD spectral tuning slightly outperforms LQ/QR. Fig.~\ref{fig.dino_clip4} visualizes the impact of different output constraint choices (no ReLU, softplus, and ReLU) on SVD spectral tuning. The results show that using ReLU achieves the best performance.

\textbf{Optimization on the Stiefel manifold.} We conduct ablations on different optimization methods on the Stiefel Manifold. We compare the original OFT, KOFT (OFT with Kronecker product), KOFT-Cayley (OFT with Kronecker product and Cayley parameterization), and OFT-Stiefel (OFT with Stiefel optimizer). From Fig.~\ref{fig:dino_clip3}, we observe that the Stiefel optimizer~\cite{kong2022momentum} outperforms the other methods when using a small learning rate, while the other methods perform similarly to each other. This demonstrates that the Stiefel optimizer can achieve comparable performance to Cayley parameterization, with the added flexibility of allowing a non-square matrix. Furthermore, the Stiefel optimizer exhibits greater robustness, as it achieves better performance with smaller learning rates compared to the other methods and achieves comparable performance with larger learning rates.

\section{Conclusion and Discussion}
\label{sec:conclusion}
In this paper, we first identify the limitations of previous PEFT methods, which are not designed to fully utilize the prior knowledge in the pre-trained weights. To address this issue, we propose spectrum-aware parameter-efficient fine-tuning, a novel approach that leverages the spectrum of the pre-trained parameters. Specifically, we introduce Spectral Orthogonal Decomposition Adaptation (SODA), which jointly performs spectral and orthogonal tuning. Experiments on diffusion personalization demonstrate that our method outperforms previous PEFT methods. Furthermore, ablation studies validate the effectiveness of the individual components of our proposed SODA approach.

\textbf{Limitations.} Our proposed fine-tuning method runs slower during training compared to LoRA due to the Stiefel optimizer. Future work will focus on accelerating the optimization algorithms and applying them to large language models.

\textbf{Acknowledgments.}
We would like to thank Haizhou Shi for valuable discussions.

{
\bibliographystyle{abbrv}
\bibliography{reference}
}

\appendix
\clearpage
\section*{\LARGE Appendix}
\markboth{Appendix}{Appendix}

\section{Derivations}
    
    


\begin{remark}
    If $\mV_1, \mV_2, \ldots, \mV_r$ are orthogonal matrices, then their Kronecker product $\bigotimes_{i=1}^r{\mV_i}=\mV_1 \otimes \mV_2 \otimes \cdots \otimes \mV_r$ is also orthogonal.
\end{remark}

\begin{proof}
    Without loss of generality, let's assume $\mV_i\in\mathbb{R}^{m_i\times n_i}$ and $\mV_i^\top\mV_i=\mI_{n_i\times n_i}$.
    
    We begin by verify $\mV=\mV_i\otimes\mV_j$ is orthogonal,
    \begin{align}
        \mV^\top\mV &= (\mV_i\otimes\mV_j)^\top(\mV_i\otimes\mV_j)\\
        &= (\mV_i^\top\otimes\mV_j^\top)(\mV_i\otimes\mV_j)\\
        &= (\mV_i^\top\mV_i)\otimes(\mV_j^\top\mV_j)\\
        &= \mI_{n_i\times n_i} \otimes \mI_{n_j\times n_j}\\
        &= \mI_{n_i n_j\times n_i n_j}
    \end{align}
    When both $\mV_i$ and $\mV_j$ are square matrices, similarly, we have $\mV\mV^\top=\mI$. Also, the determinant $|\mV|=|\mV_i|^{n_i} |\mV_j|^{n_j} = 1$.

    The final result can be obtained by iteratively applying the above.
\end{proof}

Derivation of gradient of gradients:

\noindent \textbf{Gradient of singular values.} We have $\vh=\mW\vx$ and $\mW=\mU\bSigma\mV^\top$. Then
\begin{align}
    \text{vec}(\delta\bSigma) &= \text{vec}(\delta\mW)^\top (\mV \otimes \mU)\\
    &= \text{vec}(\delta\vh\vx^\top)^\top (\mV \otimes \mU)\\
    &= (\vx\otimes\delta\vh)^\top (\mV \otimes \mU)\\
    &= (\vx^\top\mV)\otimes(\delta\vh^\top\mU)\\
    &= ((\vx^\top\mV)^\top\otimes(\delta\vh^\top\mU)^\top)^\top\\
    &= \text{vec}((\mV^\top\vx\delta\vh^\top\mU)^\top)\\
    &= \text{vec}(\mU^\top(\delta\vh\vx^\top)\mV).
\end{align}
Thus $\delta\bSigma=\mU^\top\delta\vh\vx^\top\mV$. Here $\text{vec}(\cdot)$ means staking the rows of a matrix.

\noindent \textbf{Frobenius norm of weight change.}
\begin{align}
    ||\Delta\mW'||_F^2 &= \text{trace}(\Delta\mW' \Delta\mW'^\top)\\
    &= \text{trace}(\mU\Delta\bSigma\mV^\top\mV\Delta\bSigma^\top\mU^\top)\\
    &= \text{trace}(\mU\Delta\bSigma\Delta\bSigma^\top\mU^\top)\\
    &= \text{trace}(\Delta\bSigma\Delta\bSigma^\top\mU^\top\mU)\\
    &= ||\Delta\bSigma||_F^2\\
    &= ||(\mU^\top \Delta\mW \mV)\odot \mI_{m \times n}||_F^2\\
    &\leq ||\mU^\top \Delta\mW \mV||_F^2\\
    &= \text{trace}(\mU^\top\Delta\mW\mV\mV^\top\Delta\mW^\top\mU)\\
    &= ||\Delta\mW||_F^2.
\end{align}

\section{Implementation Details}
\subsection{Dataset}
For Subject Personalization, we fine-tuned the SDXL model on the DreamBooth dataset \cite{ruiz2023dreambooth} following the Direct Consistency Optimization (DCO) framework \cite{lee2024direct}. The SDXL model was fine-tuned on 30 subjects, with each subject having 3-5 images and corresponding comprehensive captions generated by GPT-4 \cite{openai2024gpt4}. The model was trained with all PEFT methods for 1000 steps with a batch size of 1. For evaluation, we sampled 16 images for 10 prompts per subject. Seven of these prompts were designed to alter the texture of the original subject, such as ``a [$V$] [berry\_bowl] made of lego.'' These prompts were chosen because we observe that they present a significant challenge for personalization, requiring the fine-tuned model to learn new concepts without overfitting the subject.

\begin{figure}[t]
\centering
\includegraphics[width=1\textwidth]{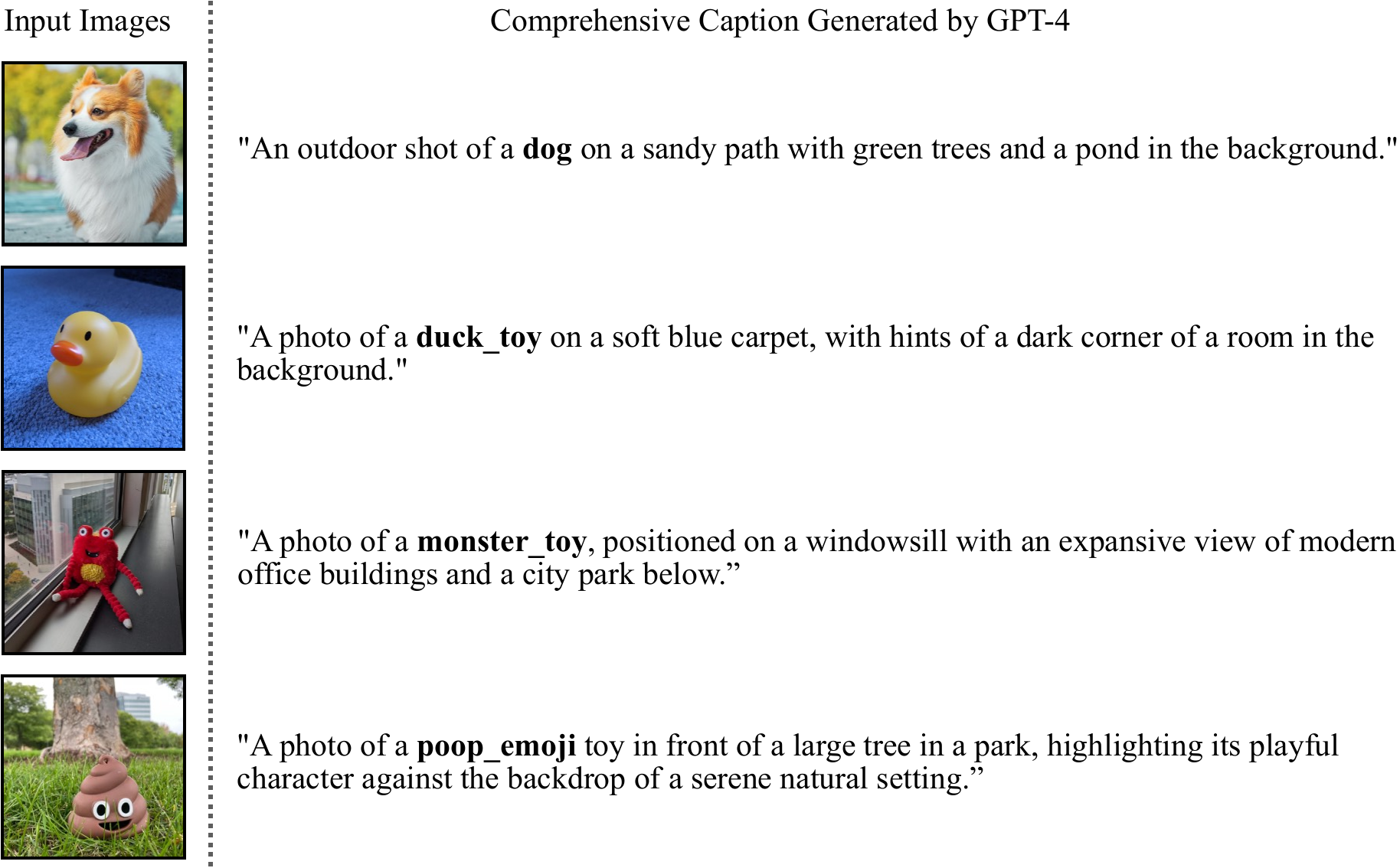} 
\caption{Results for \textbf{Examples of comprehensive captions generated by GPT-4}. The class tokens are marked in bold (e.g., dog, duck\_toy, monster\_toy).} 
\label{fig.captions} 
\end{figure}

Examples of input images and their corresponding captions generated by GPT-4 is provided in Fig. \ref{fig.captions}. And we use ``\texttt{pll}'' as the placeholder [$V$] during subject personalization (e.g., An outdoor shot of a [$V$] dog on a sandy path with green trees and a pond in the background.).

For style personalization, we fine-tuned the SDXL model on the Styledrop dataset \cite{sohn2024styledrop}, with a single reference image for 10 different styles.

\subsection{Hyperparameters}
\begin{figure}[ht]
    \centering
    \includegraphics[width=0.5\textwidth, trim=0 0 0 0, clip]{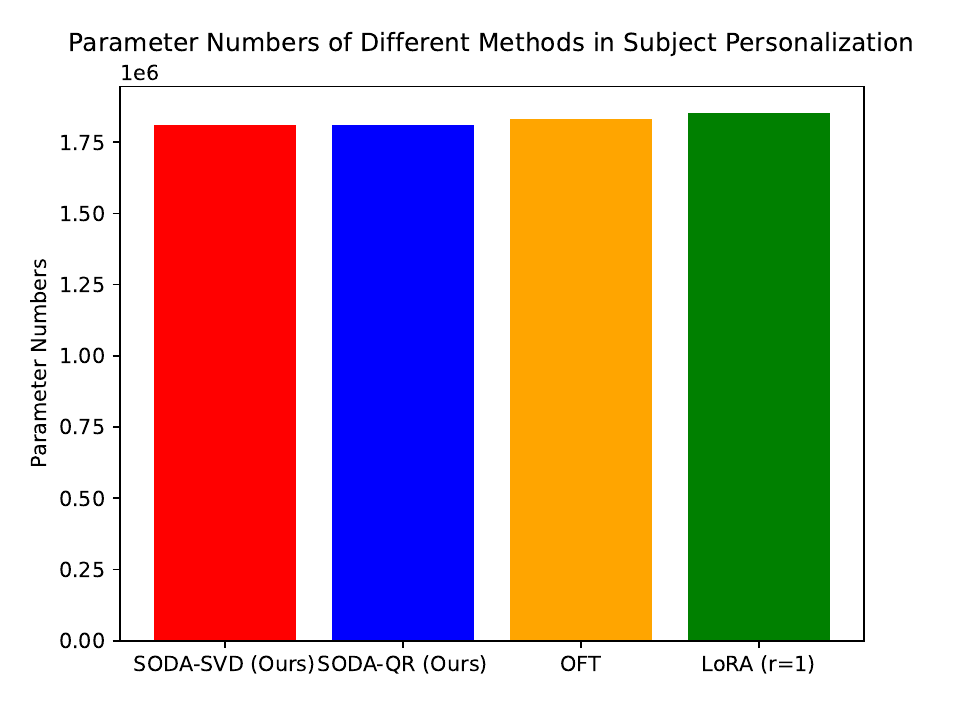}
    \caption{Parameters numbers for different methods} 
    \label{fig.params}
\end{figure}

\noindent\textbf{Budget.} We conducted all our experiments on a single GPU (e.g., A100) using a batch size of 1. We fine-tuned all the methods with 1000 optimization steps and evaluated each methods at this iteration count. To ensure fair comparisons, we hardcoded the architecture of OFT (shared orthogonal blocks) and our methods (with the orthogonal matrix constructed by the Kronecker Product of 3 small matrices). The parameter numbers for each method are detailed in Tab. \ref{fig.params}.

\begin{figure}[ht]
\centering
\includegraphics[width=1\textwidth]{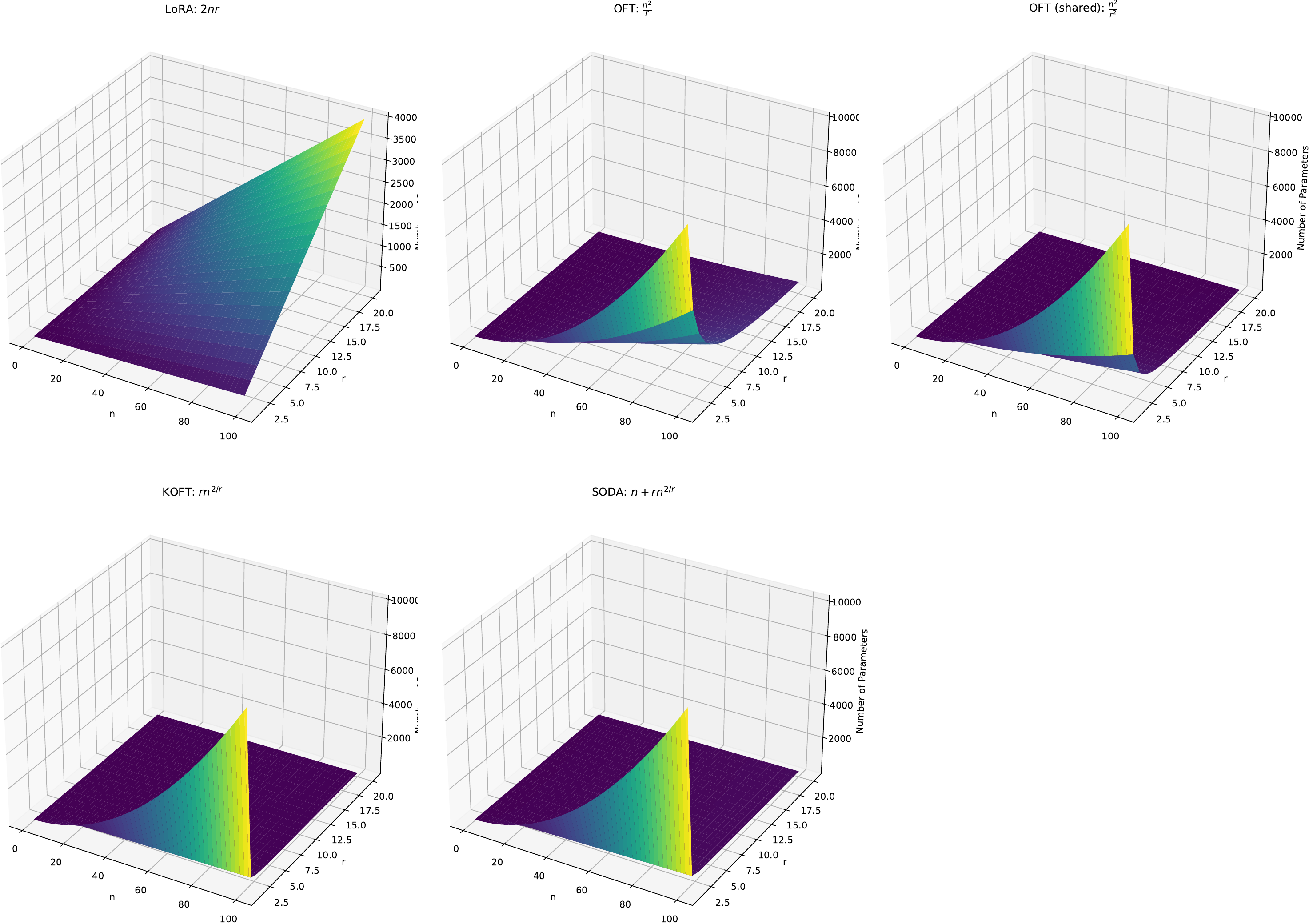} 
\caption{Number of parameters for different methods.} 
\label{fig.numparam} 
\end{figure}

\noindent\textbf{Other training settings.} With all the training without orthogonal requirements, we use the AdamW \cite{loshchilov2019decoupled} optimizer. For fair comparisons between experiments that require orthogonality, we use Adam Optimizer \cite{kingma2017adam} for OFT and Stiefel optimizer \cite{kong2022momentum} for our methods. For the DCO loss \cite{lee2024direct}, we use $\beta = 1.0$.


\section{Additional Visual Results}
\begin{figure}[ht]
\centering
\includegraphics[width=1\textwidth]{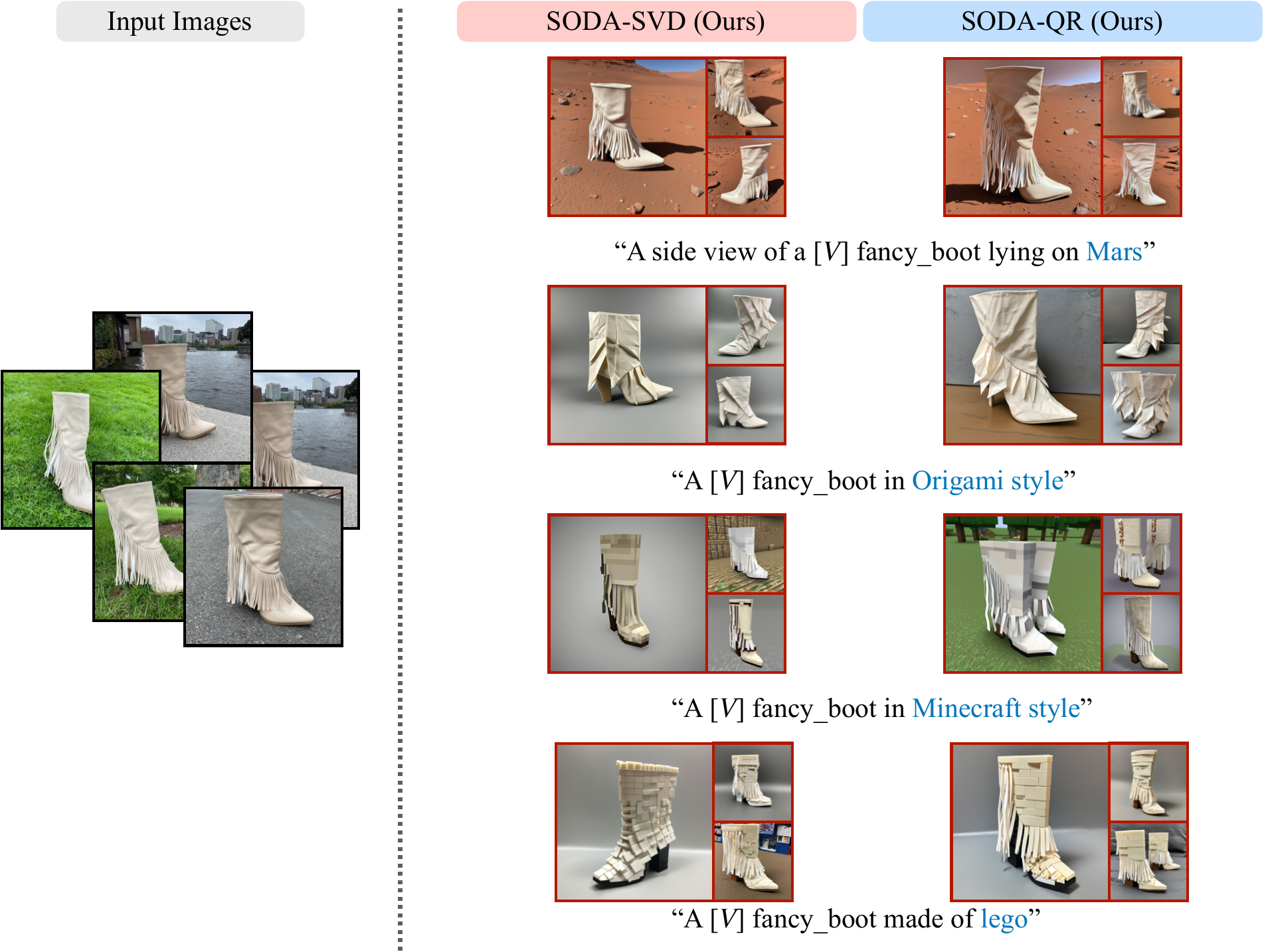} 
\caption{Results of Our Methods for the subject \textbf{fancy\_boot}.} 
\label{fig.add1} 
\end{figure}

\begin{figure}[ht]
\centering
\includegraphics[width=1\textwidth]{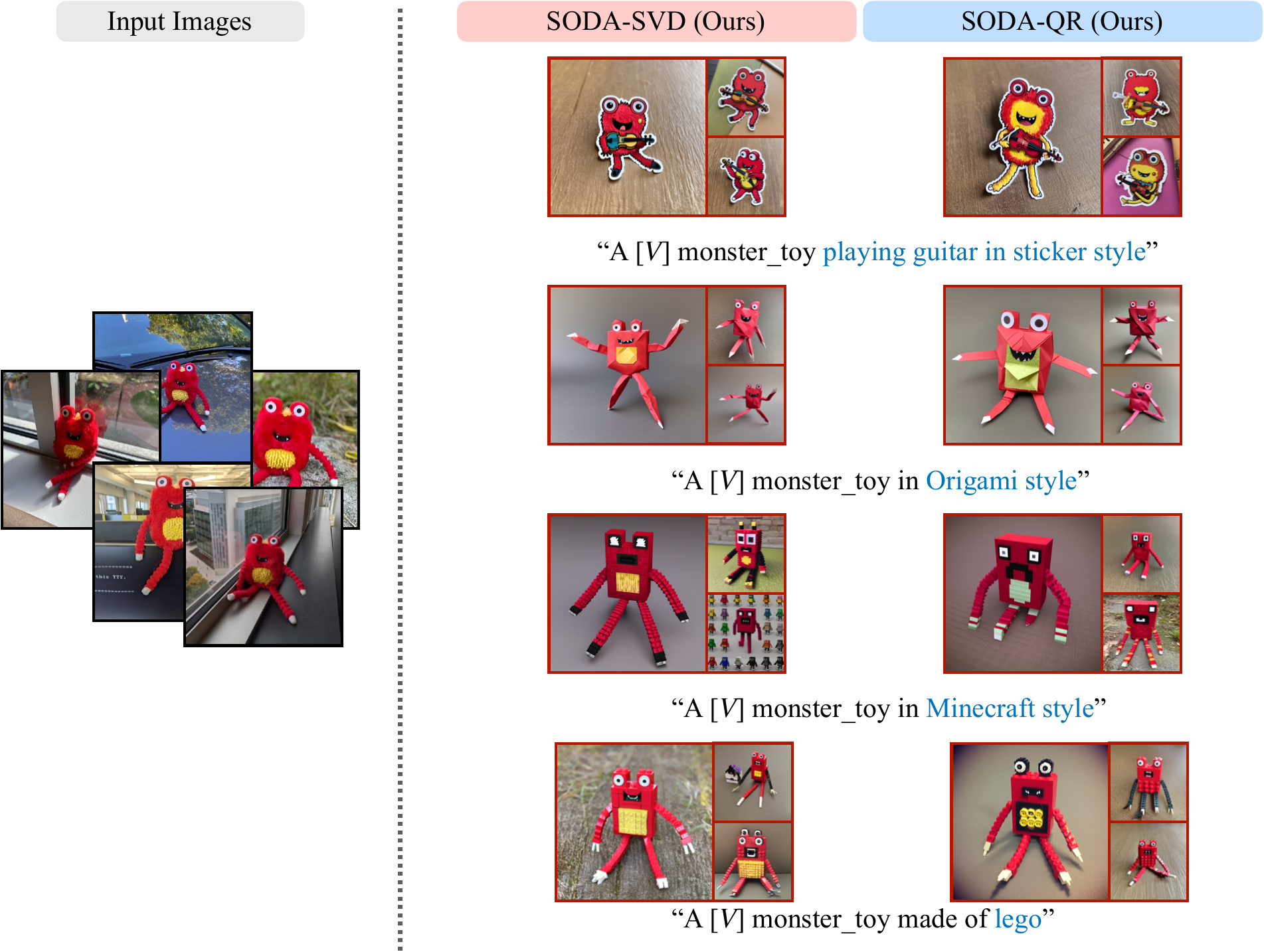} 
\caption{Results of Our Methods for the subject \textbf{monster\_toy}.} 
\label{fig.add2} 
\end{figure}

\begin{figure}[ht]
\centering
\includegraphics[width=1\textwidth]{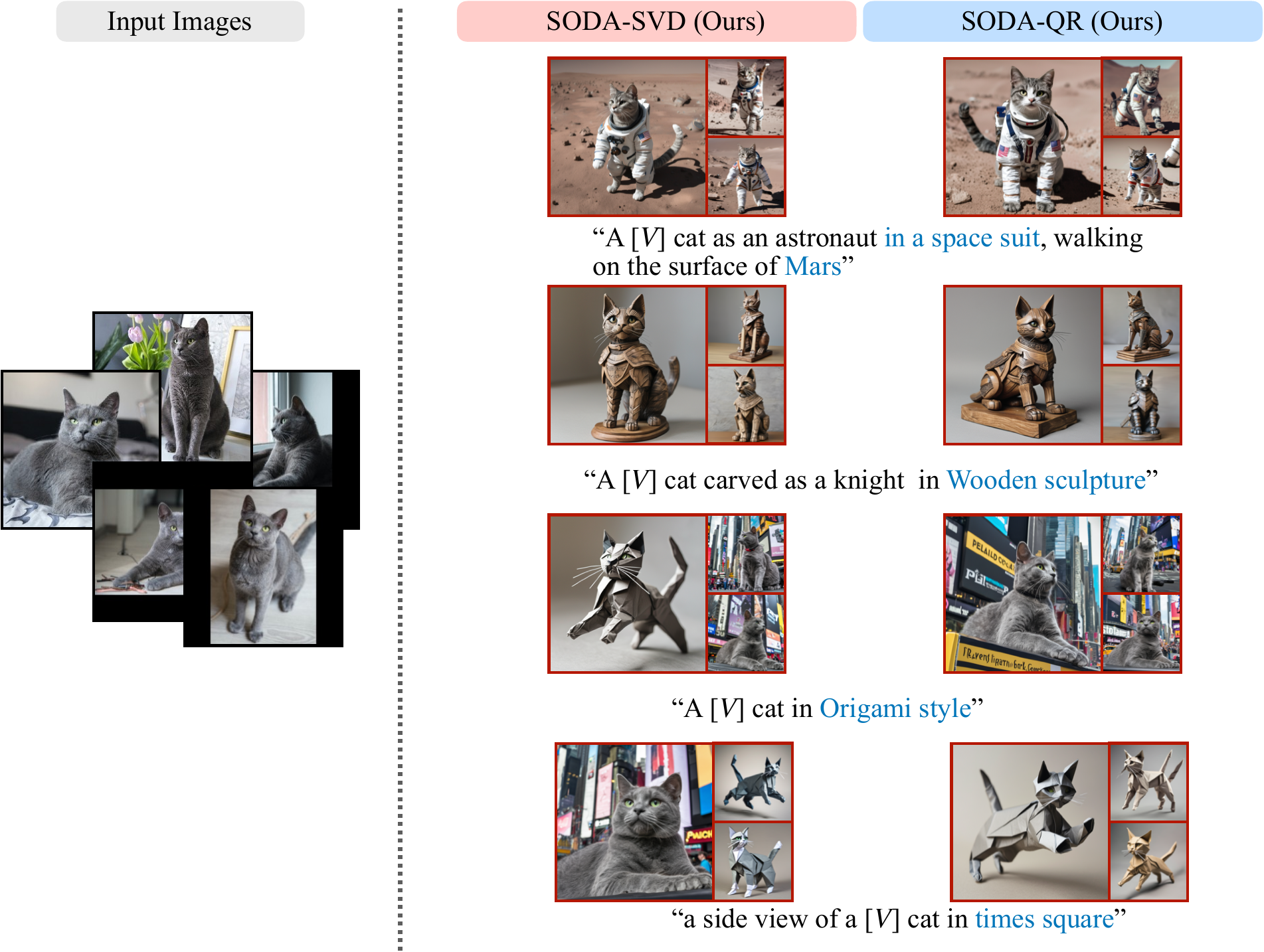} 
\caption{Results of Our Methods for the subject \textbf{cat}.} 
\label{fig.add3} 
\end{figure}

\begin{figure}[ht]
\centering
\includegraphics[width=1\textwidth]{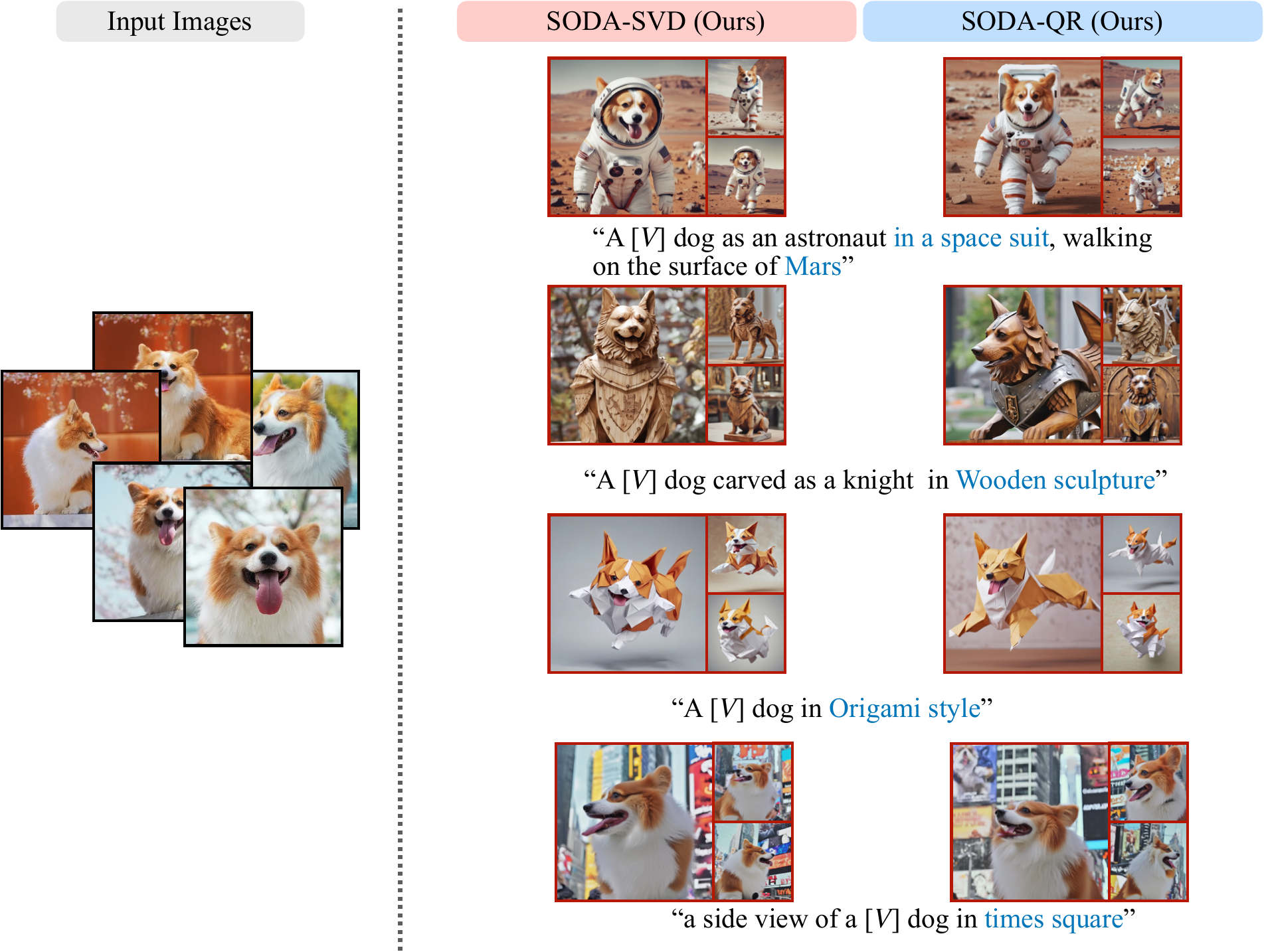} 
\caption{Results of Our Methods for the subject \textbf{dog}.} 
\label{fig.add4} 
\end{figure}

We show additional visual results in Fig.~\ref{fig.add1} \ref{fig.add2} \ref{fig.add3} \ref{fig.add4}.

\end{document}